\documentclass[conference]{IEEEtran}
\IEEEoverridecommandlockouts
\usepackage{cite}
\usepackage{amsmath,amssymb,amsfonts}
\usepackage[algo2e]{algorithm2e} 

\usepackage{graphicx}
\usepackage{textcomp}
\usepackage{xcolor}
\usepackage{comment}

\ifCLASSINFOpdf
\else
\fi
\begin{document}
\title{Large Language Model Integrated Healthcare Cyber-Physical Systems Architecture}


\author{\IEEEauthorblockN{Malithi 
Wanniarachchi Kankanamge, Syed Mhamudul Hasan, Abdur R. Shahid, Ning Yang}
\IEEEauthorblockA{\textit{Southern Illinois University, Carbondale, IL, USA} \\
malithi.mithsara@siu.edu, syedmhamudul.hasan@siu.edu, shahid@cs.siu.edu, nyang@siu.edu}
}



%


\maketitle

\begin{abstract}

Cyber-physical systems have become an essential part of the modern healthcare industry. The healthcare cyber-physical systems (HCPS) combine physical and cyber components to improve the healthcare industry. While HCPS has many advantages, it also has some drawbacks, such as a lengthy data entry process, a lack of real-time processing, and limited real-time patient visualization. To overcome these issues, this paper represents an innovative approach to integrating large language model (LLM) to enhance the efficiency of the healthcare system. By incorporating LLM at various layers, HCPS can leverage advanced AI capabilities to improve patient outcomes, advance data processing, and enhance decision-making.

\end{abstract}


\IEEEpeerreviewmaketitle

\section{Introduction}

Healthcare cyber-physical systems (HCPS) embody the convergence of cybernetics, software, and physical components, offering profound advancements in medical care~\cite{zhang2015health}. These systems, which are critical facilitators of automated healthcare procedures and data-driven medical decisions, are foundational to modern healthcare infrastructures. However, HCPS faces significant challenges, such as data fragmentation, the complexity of data interpretation, maintaining effective patient communication, manual data processing inefficiencies, and privacy concerns, which compromise system performance and patient outcomes. These issues highlight the critical need for innovative solutions that can seamlessly integrate and synthesize disparate data sources, enhance data security, and automate complex processes. Large Language Model (LLM) has the potential to transform HCPS by offering cutting-edge AI technologies. These cutting-edge LLM models, like the Generative Pre-trained Transformer (GPT) models from OpenAI, can read, understand, and produce human-like content at remarkable lengths~\cite{roumeliotis2023chatgpt}. A vast family of LLM models, including the popular GPT model, is revolutionizing Natural Language Processing (NLP). LLM uses a large training corpora of natural language, structured data, and code that are capable of processing database tuples and schemas, and plots. Thus, the wide-ranging effects of data on LLM can adapt to new industry standards and best practices. By incorporating LLM within different layers in HCPS frameworks, we can significantly streamline data collection, improve the accuracy of medical data entry, and facilitate more intuitive patient-provider interactions through natural language interfaces. This paper proposes the integration of LLM into the HCPS framework to tackle existing shortcomings and optimize healthcare delivery. Our objective is to examine how LLM can enhance decision-making processes, boost system efficiency, and ultimately lead to superior patient outcomes.



\begin{figure} 
    \centering
    \includegraphics[width = 0.49\textwidth]{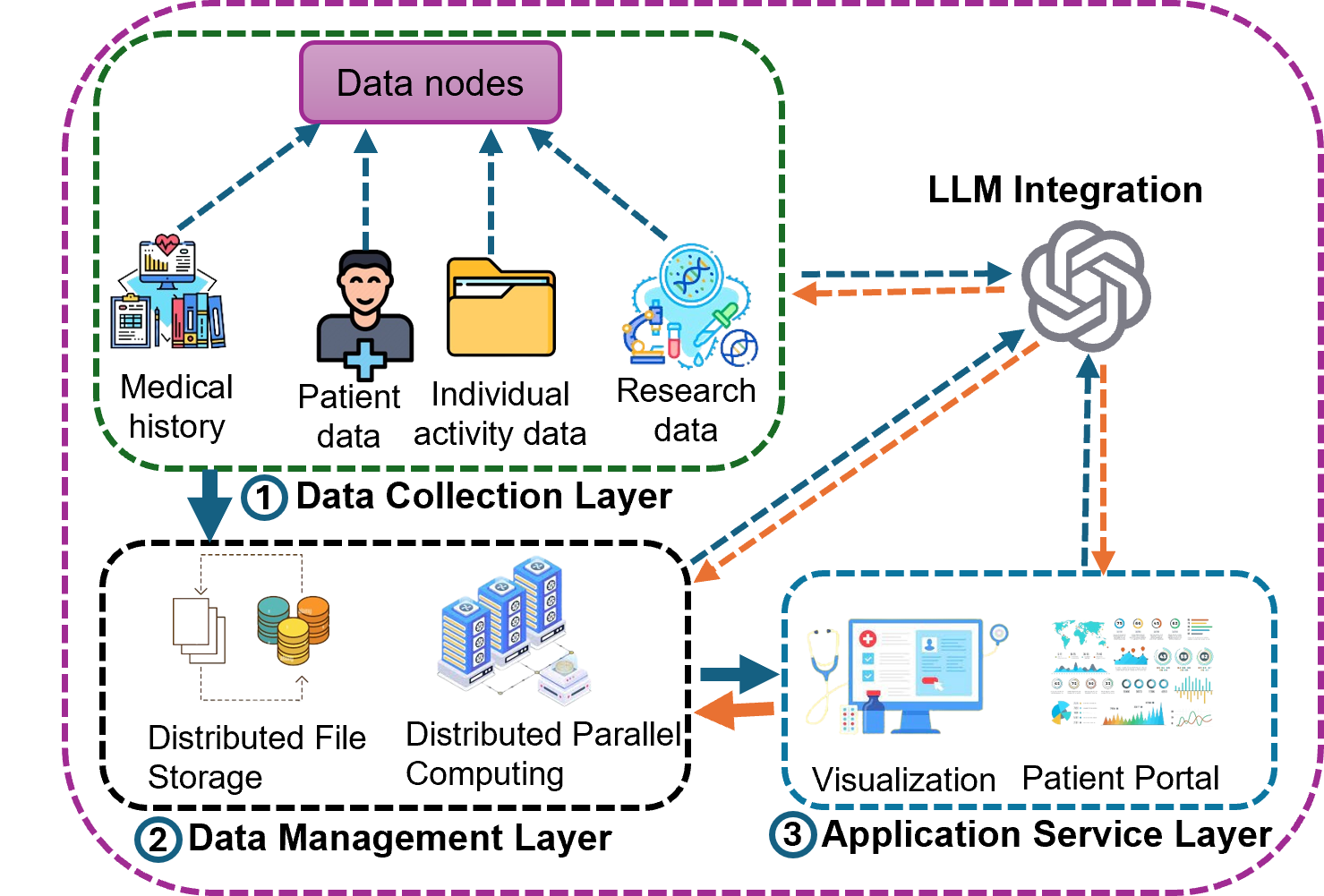}
    \vspace{-15pt}
    \caption{
The three-layered architecture of the HCPS, where the data collection layer, data management layer, and application service layer are connected to the Large Language Model (LLM).}
    \label{fig:llm}
    \vspace{-15pt}
\end{figure}

\section{System Architecture}

This study investigates the integration of LLMs into HCPS by structuring HCPS into three disctince layers: the data collection layer, the data management layer, and the application service layer~\cite{zhang2015health}, as depicted in the Figure~\ref{fig:llm}.

\subsubsection{\textbf{Data Collection Layer}}

The data collection layer plays a crucial role in acquiring multi-source, heterogeneous medical data from various resources in HCPS. An essential part of this layer is the data node. In traditional HCPS frameworks, data nodes gather information from a variety of sources, including hospitals, public health agencies, and research institutions~\cite{zhang2015health}.

LLM can include data on individual activities, clinical data, and medical history. For instance, patients can communicate with LLM through a natural language interface, entering their medical history, symptoms, and other relevant information conversationally. Therefore, the data entry process can be automated by generating electronic health records (EHRs) or other databases at the data collection layer. This reduces the possibility of human error and saves time for healthcare practitioners by eliminating the need for manual data entry. In addition, by integrating LLM into the data collection layer, patients' health status can be monitored in real-time by querying patients' symptoms, medication adherence, and other relevant information. It helps healthcare professionals identify changes or patterns in their patients health and take preventive measures through this continuous monitoring. In this layer, data privacy is imperative and for that, we need a reliable platform for developing custom LLMs. There are two options: develop a custom LLM or use third-party companies. Notable companies such as OpenAI, Google, Microsoft, etc. have trained their LLM chatbots to collect usage data, which poses a significant risk when handling sensitive information. Sharing sensitive data to third parties is dangerous, and developing LLM through third-party companies also has disadvantages because these organizations constantly invent and may change, deprecate, or modify their products without prior notice. Though custom LLM also has development and maintenance costs, using custom LLM takes less risk considering the security and privacy of medical data than using third-party LLM.

\subsubsection{\textbf{Data Management Layer}}

In the data management layer, LLM can improve the handling and analysis of medical data efficiently. In contrast to conventional data management, LLM depicts text material encoded as high-dimensional vectors of floating-point values, where each dimension corresponds to a distinct semantic aspect of the text. Rather than looking for particular symbols, query evaluation in LLM involves algebraic operations on numerical data representations. Additionally, we can use LLM as a tool for information retrieval. By posting inquiries as natural language prompts, data is retrieved as response~\cite{fernandez2023large}. For further analysis, we can divide the data management layer into two modules: distributed file storage (DFS) and distributed parallel computing (DPC) modules. The Distributed File Storage (DFS) mechanism is essential for large amounts of data in the healthcare system because it provides efficient storage, high-efficiency data upload, and versatile user management. Furthermore, the distributed parallel computing (DPC) module provides offline calculation of massive volumes of data~\cite{rashid2018distributed}. These two modules help LLM excel in enhancing data storage and real-time data processing capacity~\cite{ding2023hpc}. When an LLM-based chatbot in healthcare deals with patient data, the doctor can query information about a patient's family medical history and medication history in natural language, and the chatbot can generate an SQL command to query the entire database.

\subsubsection{\textbf{Application Service Layer}}

The traditional application service layer of HCPS includes statistical and data mining techniques to visualize patient data. We aim to incorporate LLM into this layer so that the user can see the results of the visual data analysis by displaying patient data using statistical or predictive methods through queries. For instance, a physician can ask a chatbot with an LLM to use graphs creation, which helps make decisions based on a patient record. Furthermore, this layer can improve patient engagement and data accuracy through medication reminders, information about the patient's current medications like side effects, dosage, etc., and notifying them if there are any medication conflicts. Additionally, it is helpful to have services like chatbots for appointment scheduling to establish new appointments as well as edit existing ones. We can also use the LLM to resolve scheduling conflicts as needed. In this layer, we introduce the patient portal integrating LLM that allows patients and healthcare providers to access their medical information. Finally, this layer's integration of LLM will provide a real-time interface for patients and healthcare professionals to make decisions.

\section{Realization Challenges}

The realization of an LLM-integrated HCPS architecture presents several challenges that must be meticulously addressed to ensure the successful deployment and operation of such systems. Technical challenges include ensuring compatibility with existing systems, managing scalability, and achieving interoperability across diverse healthcare systems and applications. Privacy and security are paramount, with a need for robust data protection measures, compliance with stringent healthcare regulations, and effective anonymization techniques to safeguard patient privacy. Ethical considerations such as preventing bias in AI algorithms, maintaining transparency in AI decision-making, and preserving the integrity of the patient-provider relationship are critical. Additionally, the sustainability of deploying such advanced technologies involves assessing the environmental impact of the massive data centers required to train and run LLMs, promoting energy-efficient computing practices, and considering the lifecycle management of these systems\cite{hasan2024towards}. Furthermore, economic and operational challenges, such as cost justification, professional training, and continuous system updates, are critical for the effective integration of LLM in HCPS.

\section{Conclusion}

This paper introduces a novel framework for healthcare systems that incorporates LLM. Although LLM can have some potential misuse, the overall performance of medical applications increases when LLMs are integrated into data collection, data management, and application service layers. Addressing challenges like security, privacy, cost, and integration, the LLM can revolutionize the current HCPS.

\bibliographystyle{IEEEtran}
\bibliography{main}

\end{document}